\documentclass[letterpaper, preprint, paper,11pt]{AAS}	

\usepackage{bm}
\usepackage{amsmath}
\usepackage{subfigure}
\usepackage[colorlinks=true, pdfstartview=FitV, linkcolor=black, citecolor= black, urlcolor= black]{hyperref}
\usepackage{overcite}
\usepackage{footnpag}			      	
\usepackage{multirow}

\usepackage{siunitx}

\PaperNumber{22-813}

\begin{document}

\title{RGB-D Robotic Pose Estimation For a Servicing Robotic Arm}

\author{Jared Herron\thanks{Undergraduate, Aerospace Engineering Department, Embry-Riddle Aeronautical University, 1 Aerospace Blvd., Daytona Beach, FL, 32114.},
Daniel Lopez\footnotemark[1],
Jarred Jordan\footnotemark[1],
Jillian Rudy\footnotemark[1],
Aryslan Malik\thanks{Visiting Professor, Aerospace Engineering Department, Embry-Riddle Aeronautical University, 1 Aerospace Blvd., Daytona Beach, FL, 32114.},
Daniel Posada\thanks{Ph.D. Candidate, Aerospace Engineering Department, Embry-Riddle Aeronautical University, 1 Aerospace Blvd., Daytona Beach, FL, 32114.},
Mehran Andalibi\thanks{Associate Professor, Mechanical Engineering Department, Embry-Riddle Aeronautical University, 3700 Willow Creek Rd, Prescott, AZ, 86301.}, \ and 
Troy Henderson\thanks{Associate Professor, Aerospace Engineering Department, Embry-Riddle Aeronautical University, 1 Aerospace Blvd., Daytona Beach, FL, 32114.}
}

\maketitle{}

\begin{abstract}
A large number of robotic and human-assisted missions to the Moon and Mars are forecast. NASA's efforts to learn about the geology and makeup of these celestial bodies rely heavily on the use of robotic arms. The safety and redundancy aspects will be crucial when humans will be working alongside the robotic explorers. Additionally, robotic arms are crucial to satellite servicing and planned orbit debris mitigation missions. The goal of this work is to create a custom Computer Vision (CV) based Artificial Neural Network (ANN) that would be able to rapidly identify the posture of a 7 Degree of Freedom (DoF) robotic arm from a single (RGB-D) image - just like humans can easily identify if an arm is pointing in some general direction. The Sawyer robotic arm is used for developing and training this intelligent algorithm. Since Sawyer's joint space spans 7 dimensions, it is an insurmountable task to cover the entire joint configuration space. In this work, orthogonal arrays are used, similar to the Taguchi method, to efficiently span the joint space with the minimal number of training images. This ``optimally'' generated database is used to train the custom ANN and its degree of accuracy is on average equal to twice the smallest joint displacement step used for database generation. A pre-trained ANN will be useful for estimating the postures of robotic manipulators used on space stations, spacecraft, and rovers as an auxiliary tool or for contingency plans.    
\end{abstract}

\section{Introduction}
Estimating the pose of a multi-degree of freedom robotic manipulator from a single RGB image can be considered as one of the most challenging problems in CV. It entails solving two naturally ambiguous tasks. First, the 2D locations of the manipulator's joints, or landmarks, must be determined from the image. This is a difficult task because of the wide differences in visual appearance brought on by various camera angles, self- and external occlusions, and illumination. Secondly, the space of potential joint space poses consistent with the 2D landmark locations of a manipulator is infinite, making it impossible to map the coordinates of the 2D landmarks into joint space vector from a single RGB photograph.

The main idea of this work is to develop an algorithm capable of accurately estimating the pose of a robotic system from a single RGB-D image, so that the Depth layer will be assisting with correctly mapping 2D landmarks to joint space. Applications of this algorithm could be: safety - i.e. using visual feedback to stop a robotic arm system if its joint encoders are malfunctioning; analysis - i.e. during space debris removal to identify the attitude of the debris that being removed; redundancy - the ANN would serve as an auxiliary measurement system, etc. In this work, the ANN model is trained and validated on the images taken from the same camera position and angle. Furthermore, the joint limits are constricted to a smaller range. 

The main contribution of this work is an ``optimal'' ANN architecture, which can be later re-trained and used for other robotic systems, or more generally, any object that rotates about some axis. Although a more elegant approach to manipulator pose estimation may still lie in exclusive application of CV algorithms, designing a custom ANN can be further generalized to different conditions arising from the ability of space servicing arms to detach either end of the arm releasing it from the station and reattach somewhere else, allowing it ``crawl'' along the outside of the station (e.g. Canadarm2, Canadarm3, European Robotic Arm, etc.), which lead the authors to believe that exercising the idea of custom ANN tailored for robotic systems has a substantial academic value. It is also important to mention that the custom ANN is tested on the Sawyer robotic arm which is a redundant manipulator with no unique inverse kinematics solution.

\section{Demonstration Robotic system}

The commercially available 7DOF Sawyer robotic arm is controlled through Ubuntu-ROS framework while in Software Development Kit (SDK) mode in order to produce a database of RGB-D imagery of robotic arm assuming various poses. The arm can be seen in Figure \ref{fig:sawyer} below. 
\begin{figure}[h]
    \centering
    \includegraphics[scale=0.14]{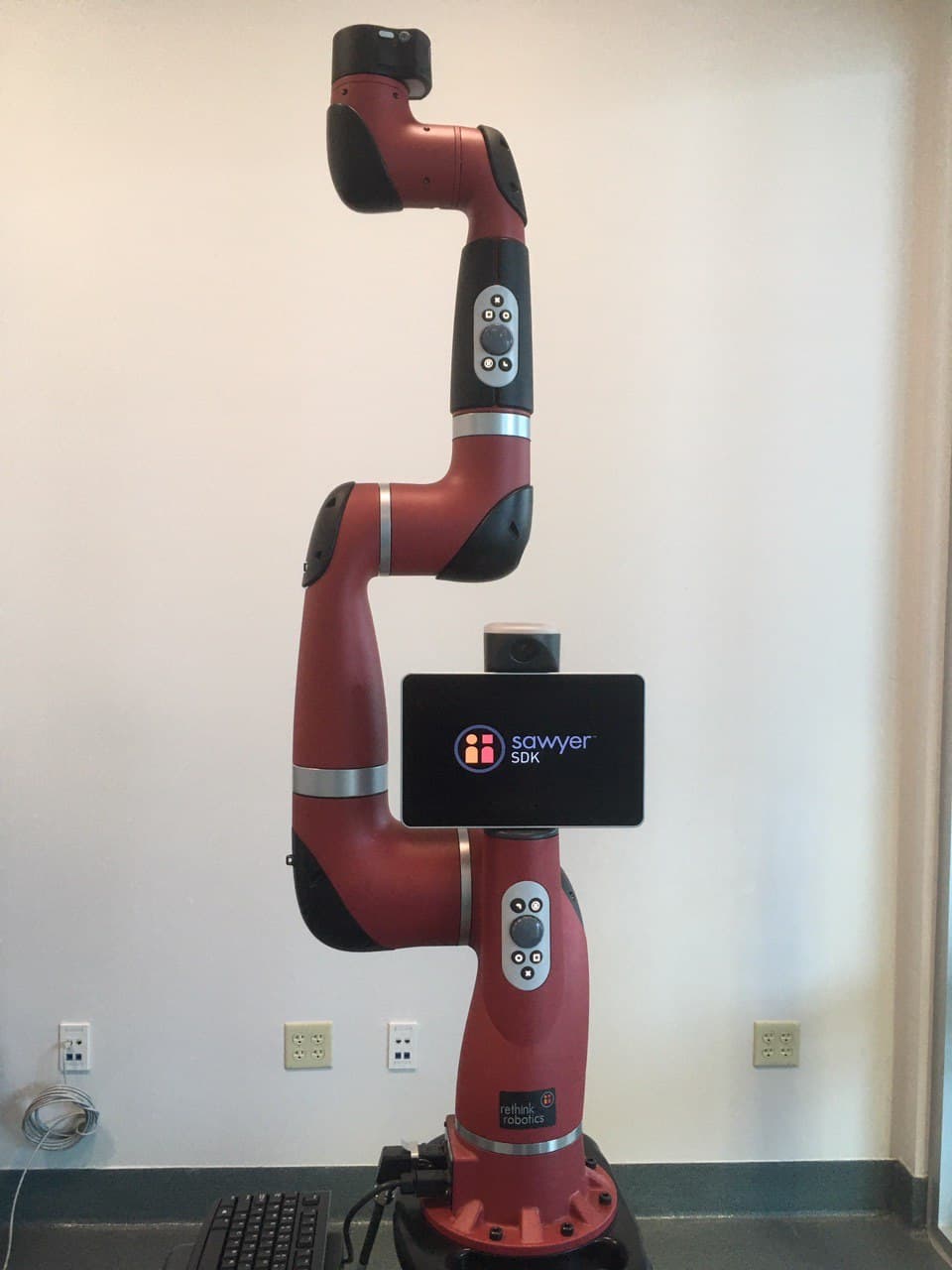}
    \caption{Sawyer robotic manipulator \cite{poe_robotics3}}
    \label{fig:sawyer}
\end{figure}

Inference models can be augmented by using Product of Exponentials (PoE) Forward Kinematics (FK) given below in Equation \ref{eq:PoE} \cite{poe_robotics1,poe_robotics2,poe_robotics3,poe_robotics4,poe_robotics5,poe_robotics6,malik2022using}:

\begin{equation}\label{eq:PoE}
      {T_{sn}}=e^{[\mathcal{S}_1]\theta_1}e^{[\mathcal{S}_2]\theta_2}\cdots e^{[\mathcal{S}_7]\theta_7}M_{s7}\in{}SE(3)
\end{equation}
where $(\theta_1,\theta_2,\dots,\theta_7)$ are joint angles with respect to the home configuration.

Unlike the conventional Denavit-Hartenberg parameters approach, the PoE does not follow strict rules as to how assign frames, it has an intuitive interpretation based in geometric mechanics, and it treats revolute and prismatic joints uniformly with a concise and elegant formula \cite{poe,lynch2017modern}.

\section{Computer Vision}
RGB-D imagery served as training, verification, and validation databases for the ANN model. OpenCV is used for structuring, manipulating, and augmenting the images to increase the number of samples within databases \cite{pulli2012real}. Any off-the-shelf RGB-D camera can be used to generate this database as the ANN model does not require high resolution images. Kinect V1/V2 \cite{wasenmuller2016comparison}, OAK-D-Lite \cite{mahendran2021computer}, and even iPhone's back facing camera satisfy the database requirements \cite{baruch2021arkitscenes}. A depth image produced using iPhone camera is shown below in Figure \ref{fig:comparison}.

\begin{figure}[h!]
    \centering
    \includegraphics[trim={0 9.3cm 0 0},clip, scale=0.65]{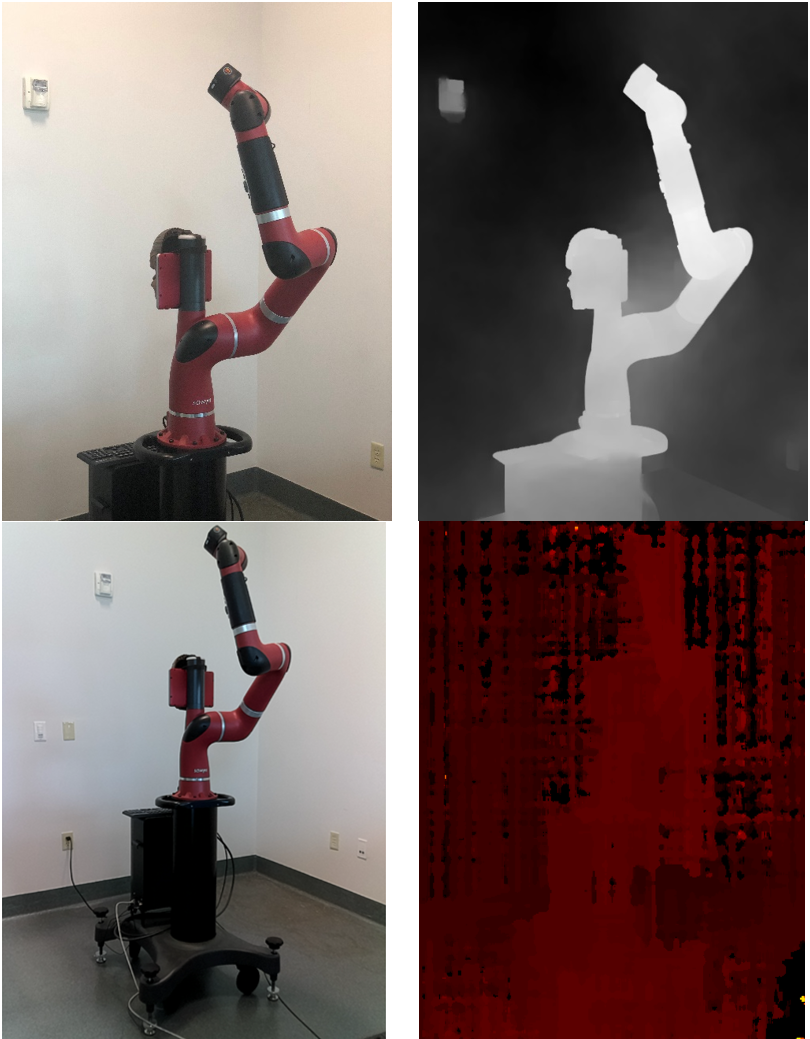}
    \caption{Top row is RGB and Depth images taken on iPhone X camera}
    \label{fig:comparison}
\end{figure}

\begin{figure}[h!]
    \centering
    \includegraphics[scale=0.65]{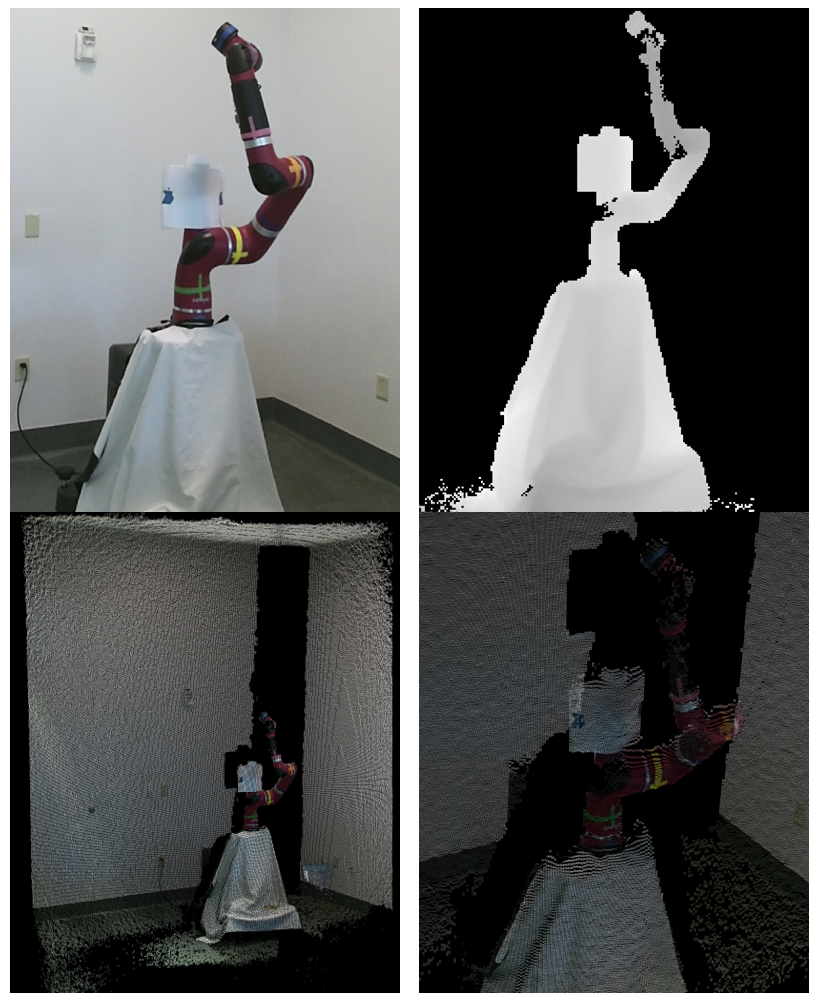}
    \caption{Top row is RGB and Depth images taken on Kinect V2, bottom row is pointclouds from two different points of view}
    \label{fig:kinectv2}
\end{figure}

After experimentation with several cameras (iPhone X, OAK-D-Lite, Intel RealSense L515, Kinect V1, Kinect V2), it was decided to use Kinect V2 as it produced the most detailed depth map out of all tested cameras. RGB, depth, and corresponding pointclouds produced using Kinect V2 are shown in Figure \ref{fig:kinectv2}.

\section{Network Architecture}
The final aspect of the work is designing the appropriate ANN architecture for the best performance and accuracy of pose estimation. Different detection and classification models exist that run an inference graph to obtain a human pose from an image \cite{tome2017lifting,wang2020deep,yu2021lite}. Developing ANN model tailored to a robotic arm pose estimation is a highly iterative task as there are numerous hyperparameters that affect the training time and the algorithm's accuracy. Furthermore, seven joints and their limits will result in a huge number of images if joint step size is chosen to be small. Thus, in this work, orthogonal arrays are employed in order to optimize the aspect of database generation. The major contribution of this approach lies in efficient (orthogonal) span of the joint space with less number of poses (joint combinations). 

The data structure of the input and output was the first consideration when developing the custom ANN. To help keep the model as general as possible, the input was chosen to be a plain RGB-D image; an image with 4 channels: Red, Green, Blue, and Depth (distance from the camera to the point captured in the pixel). The output was chosen to be the joint space of the robotic arm (7 angles corresponding to each joint). This defined the ANN to be a regression model.

Based on the choices of input and output of the model, the ANN architecture was defined by comparing the problem and proposed model to a similar problem. The most similar problem is bounding box regression, which takes an image as input and outputs a quasi-continuous output of the top left-bottom right coordinates of a bounding box \cite{he2019bounding,zheng2020distance,rezatofighi2019generalized}. Similarly, this model takes an image and should output 7 numbers corresponding to the predicted angle of each joint on the Sawyer robotic arm.

The model architecture (modeled after the bounding box problem) consists of 2 main sections: repeated 2D convolution and max pooling layers over the image, and several dense layers with the hyperbolic tangent $(\tanh)$ activation function. Hyperbolic tangent is a convenient choice for the selected constrained dataset described in Database section. If a bigger joint positions range is considered, the angles should be scaled down to for each joint, for example if first joint's limits are $(-4\pi\,\si{rad},4\pi\,\si{rad})$ a mapping to $(-1,1)$ will be required.   
After the data-set was made, the hyper-parameters were tweaked until promising results were found, this is described in greater detail in the ``Training'' section below. 

\begin{figure}[h!]
    \centering
    \includegraphics[width=\textwidth]{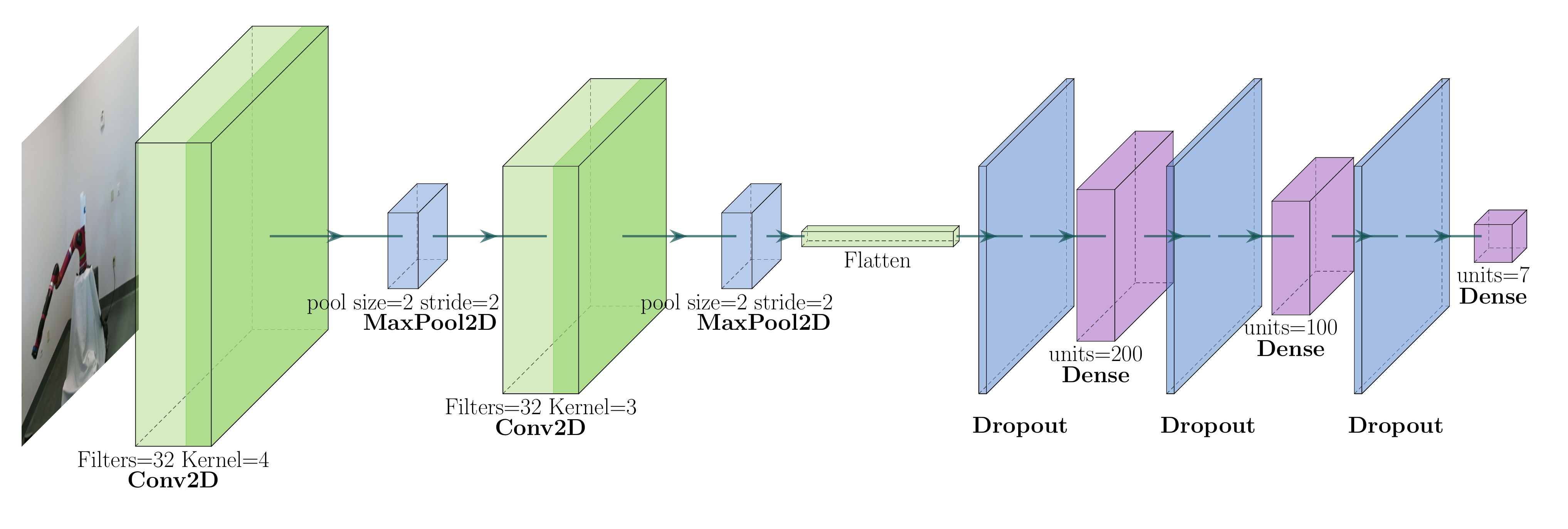}
    \caption{Neural network architecture graph}
    \label{fig:NNgraph}
\end{figure}

The final ANN architecture is comprised of eleven layers, two convolution and pooling, three dropout, three dense, and one flatten layer. The network's first layer is a convolution (2D) with kernel size of four and thirty-two filters. This allows the neural network to create a feature map, which then goes through a max pooling layer. The max pooling layer emphasizes and retains the most important features, such as the outline of the robotic arm. The max pooling layer has a stride and pool size of two. This then repeats for a second convolution (2D) layer with thirty-two filters and a kernel size of three. The kernel size is decreased in order to avoid over simplification of the image. The output is once again processed through a max pooling layer with a stride and pool size of two. After the dataset is processed by the first four layers, it is directed into a flatten layer to turn it into a one dimensional array. After flattening, the dataset is put into a dropout layer, which randomly selects array values and replaces them with zeros. This is accomplished to prevent overfitting the data. The dataset is then processed by a dense layer with 200 neurons, which is followed by another dropout layer directly after. Then, another dense layer is used, this time with 100 neurons. Another dropout layer is used, followed by a final dense layer with 7 neurons which corresponds to the estimated joint space vector. The ANN architecture is visualized in Figure \ref{fig:NNgraph}.
\section{Database}
Creating a database of RGB-D images that spans 7 dimensional joint space is a challenging task. As a first simplification, the scope of the joint space was limited to $(-55^\circ,55^\circ)$. Even with this simplification if the step is assumed to be $10^\circ$, the complete database would require taking $11^7$ (or $\approx20$ million) images. In order to efficiently span the joint space of the manipulator an orthogonal array was built using Equation \ref{eq:orthogonal_array}, which results in only $144$ images. 
\begin{equation}\label{eq:orthogonal_array}
    {OA}(N,k,s,t)\xrightarrow{}{OA}(144,7,12,2)
\end{equation}
where, $N$, $k$, $s$, $t$ denote number of runs, factors, levels, and strength respectively. This method is similar to design of experiments via Taguchi methods \cite{zhang2007orthogonal,hedayat1999orthogonal,roy2010primer}. The resulting orthogonal array is 7 by 144 matrix, which represents 144 different poses of the robotic arm. In total there are $s=12$ levels ranging from $0$ to $11$. In order to convert the orthogonal array to joint space the following equation was used:
\begin{equation}\label{eq:ordeg}
OA_{deg}=10\cdot{OA}(144,7,12,2)-55^\circ    
\end{equation}
where, $OA_{deg}$ is 144 by 7 matrix with joint angles in degrees ranging from $-55^\circ$ to $55^\circ$. The number of factors $k$ in this case is equal to the number of joints. Now, the number of runs, levels, and strength can be arbitrarily chosen. However, as the number of levels increases the number of runs will increase as well, and it is expected that the ANN inference accuracy will be improved as a result. The database generation process was as follows: Sawyer robotic arm assumes a posture from orthogonal array given in Equation \ref{eq:ordeg}, and, consequently, an RGB-D image is taken from the same point of view.

\begin{figure}[!h]
    \centering
    \begin{subfigure}
        \centering
        \includegraphics[scale=0.05,angle =-90]{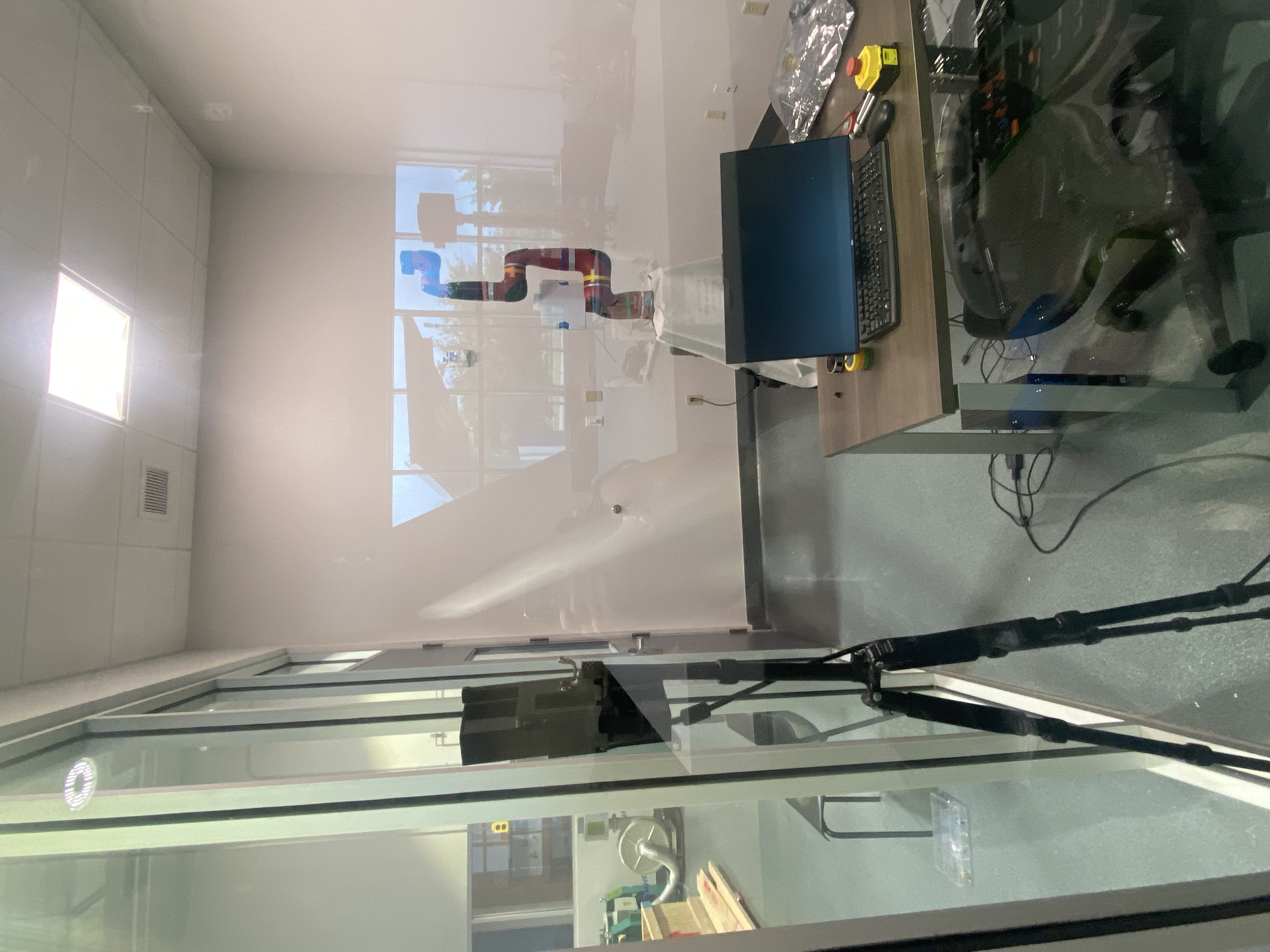}
    \end{subfigure}
    \begin{subfigure}
        \centering
        \includegraphics[scale=0.05,angle =-90]{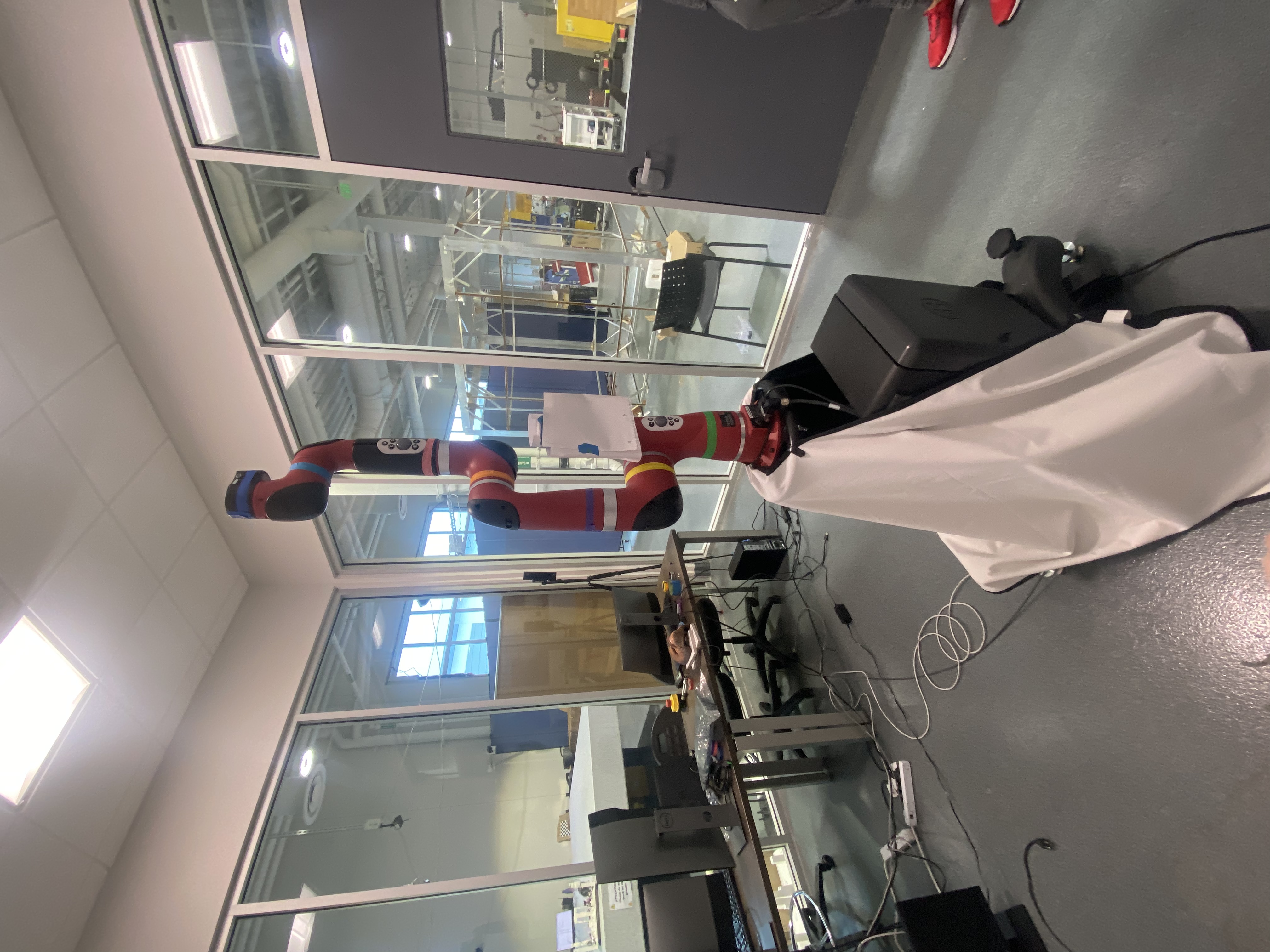}
    \end{subfigure}
    \caption{The setup from two different angles.}
    \label{fig:Kinect}
\end{figure}

The database was created using a Kinect V2 camera for both RGB and depth. The camera was set on a tripod that was set steadily in the corner of the room (for the duration of collection of training data) as shown in Figure \ref{fig:Kinect}. The robot was also given a few customizations to help the convolutional ANN with training process. The pedestal of the robot was covered with a white tarp (roughly matching the wall) to prevent confusion, as the robot has the same black color as the pedestal. Colored tape was also put on each individual joint to help the ANN in determining where each landmark (joint) is in each image. The Kinect images were then normalized as a pre-processing step before feeding them as an input to the ANN. 

A database of purely random positions was also taken using the same set up as the orthogonal array database. The random arrays were generated and the robotic arm was set to do the same amount (144) of different poses to compare the effectiveness of the orthogonal array data set. The same camera and joint angle limits were used for consistency.

The validation database was the final dataset that was generated using exactly the same process and it consisted of 32 arbitrary poses within the constrained joint space limits. Both orthogonal and random ANN models were validated using this ``out-of-sample'' database.
\section{Training}
The parameters in the neural network are incredibly important for training the model. There are many parameters that can be changed to optimize training process, but a few in particular caused drastic changes. Some notable hyperparameters were the batchsize and the optimizer function. Batchsize was originally set to 144, so the entire training database is processed by the neural network at once per epoch. The batchsize was changed to factors of 144, such as 16, 9, etc. Following an iterative search, 144 was determined to be the optimal batchsize for this ANN. The optimizing function used originally was Adam, as it is well suited for large multi-dimensional problems. The optimizer was changed to various other built-in optimization functions through TensorFlow, and it was concluded that Adam worked the best for this ANN architecture.

\begin{figure}[!h]
    \centering
    \includegraphics[scale=0.6]{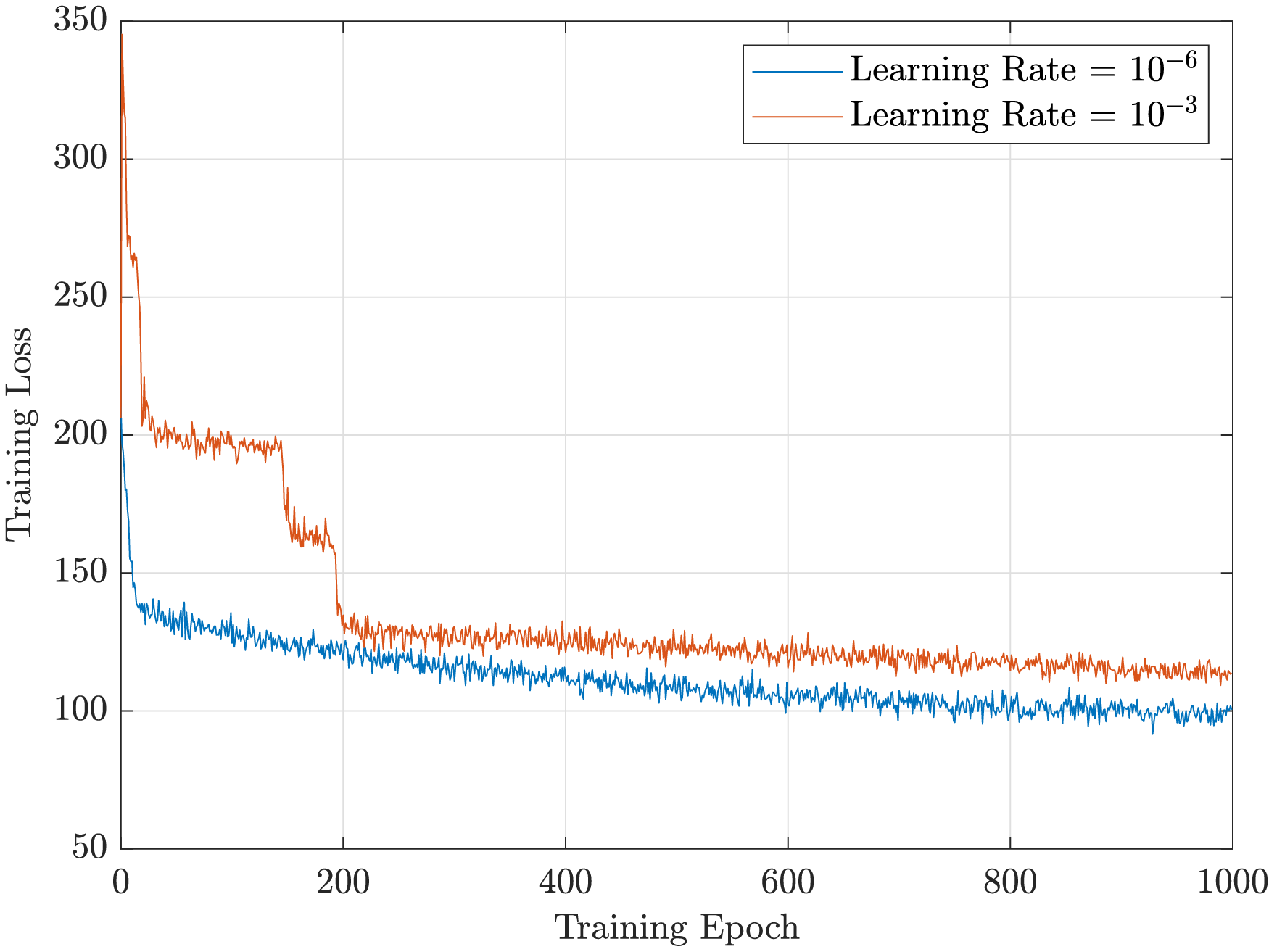}
    \caption{Blue graph shows the loss when the learning rate is decreased to $10^{-6}$. Red graph shows the loss when the learning rate is increased to $10^{-3}$. All other parameters are kept constant.}
    \label{fig:LR}
\end{figure}

Changes in the learning rate heavily affected the ANN's predictions. The learning rate defines the degree to which to adjust the model each time the model weights are updated in response to the estimated error - a learning rate that is too small could lead to a lengthy training process that could become stuck, while a value that is too large could lead to learning a suboptimal set of weights too quickly or to an unstable training process. Up to a certain point, increasing the learning rate produces more accurate pose estimates. The results of ANN become underfit if the learning rate is chosen to be too high. On the other hand, if the learning rate is too low the model overfits the estimates.\cite{LeslieHyp} Figure \ref{fig:LR} demonstrates two different training runs with vastly different learning rates. 

After numerous experimentation it was observed that anything higher than $10^{-6}$ makes the model underfit, while anything lower overfits the pose estimates. Thus, the learning rate used for the final ANN architecture is $10^{-6}$, as it demonstrated good results when ANN is  trained on orthogonal based dataset. 

\begin{figure}[!h]
    \centering
    \includegraphics[scale=0.6]{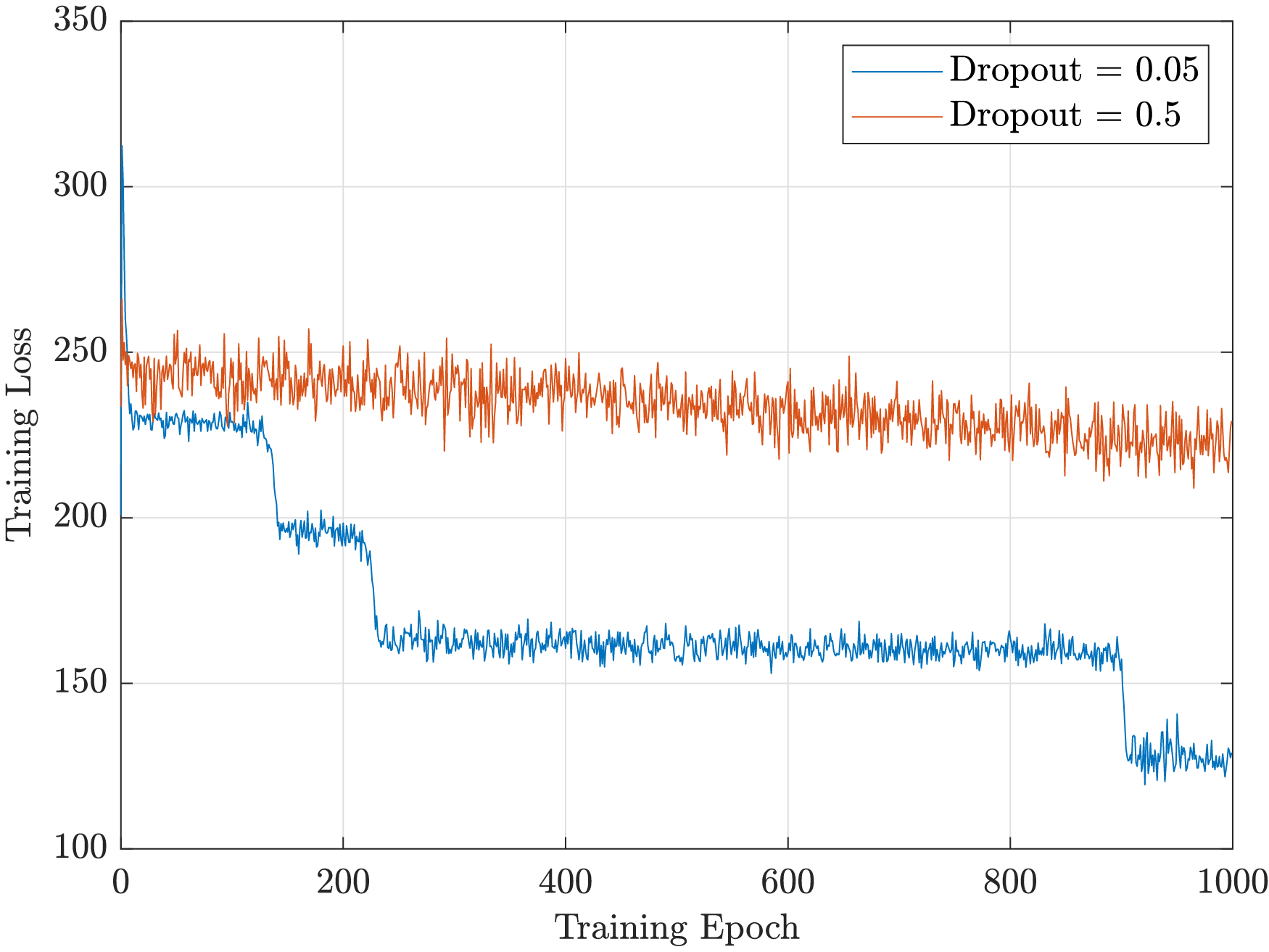}
    \caption{Red graph shows the loss when the dropout is increased to 0.5. Blue graph shows the loss when the dropout is decreased to 0.05. All other parameters are kept constant.}
    \label{fig:Dropout}
\end{figure}

The most notable change observed in the model was a change in dropout layers. Dropout causes the neural network to avoid using some of its layers at random in each epoch.\cite{LeslieHyp} The dropout heavily affected whether the model would be overfitting or underfitting. Since the dataset tasked to the ANN was so complex, a lower dropout was desired to increase accuracy in the predictions given. Higher values in dropout caused convergence failures. Evidence supporting this is shown in Figure \ref{fig:Dropout}, where dropout rates of 0.05 and 0.5 are compared.

\begin{figure}[!h]
    \centering
    \begin{subfigure}
        \centering
        \includegraphics[width=0.4\textwidth]{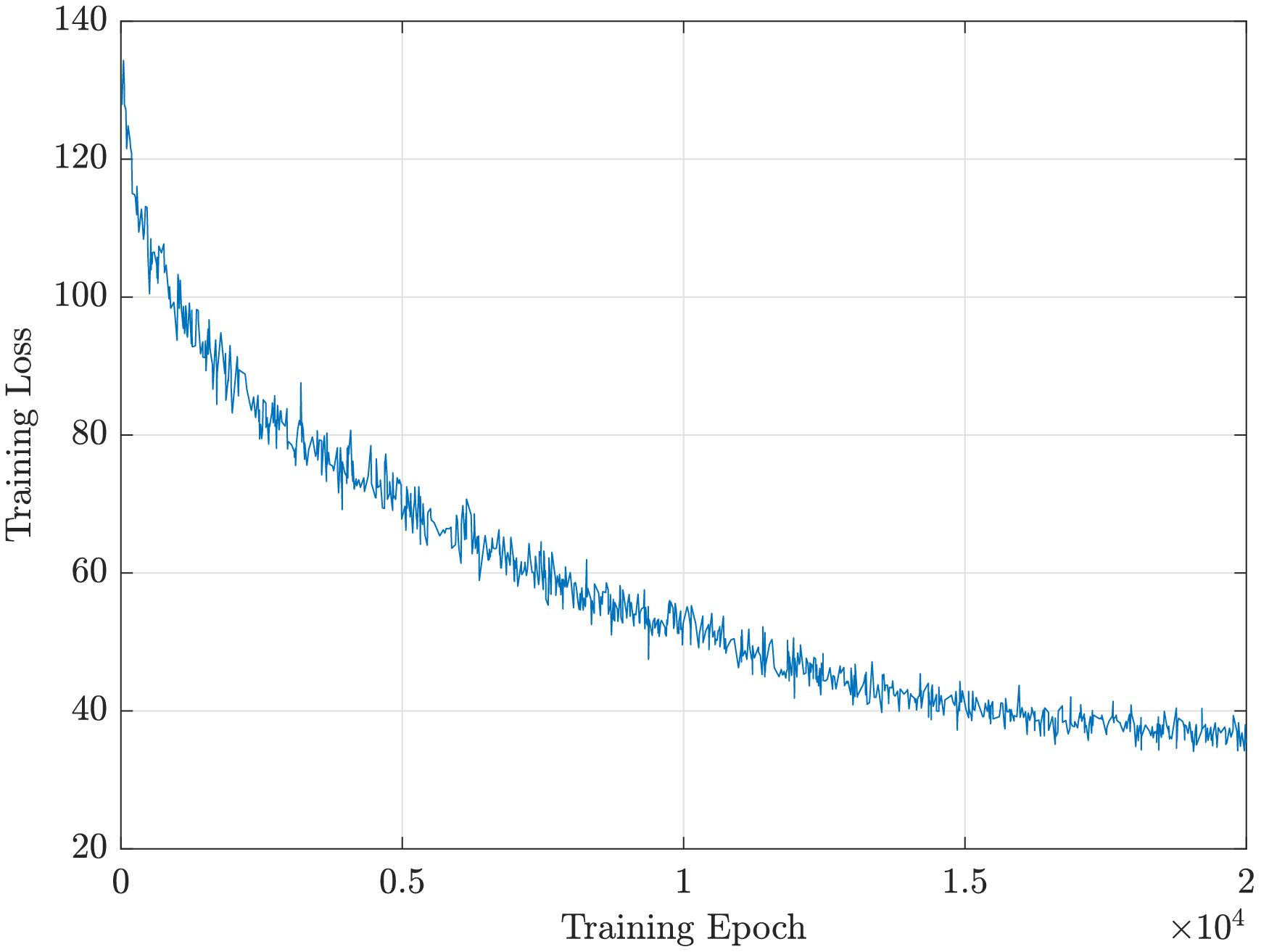}
    \end{subfigure}
    \begin{subfigure}
        \centering
        \includegraphics[width=0.4\textwidth]{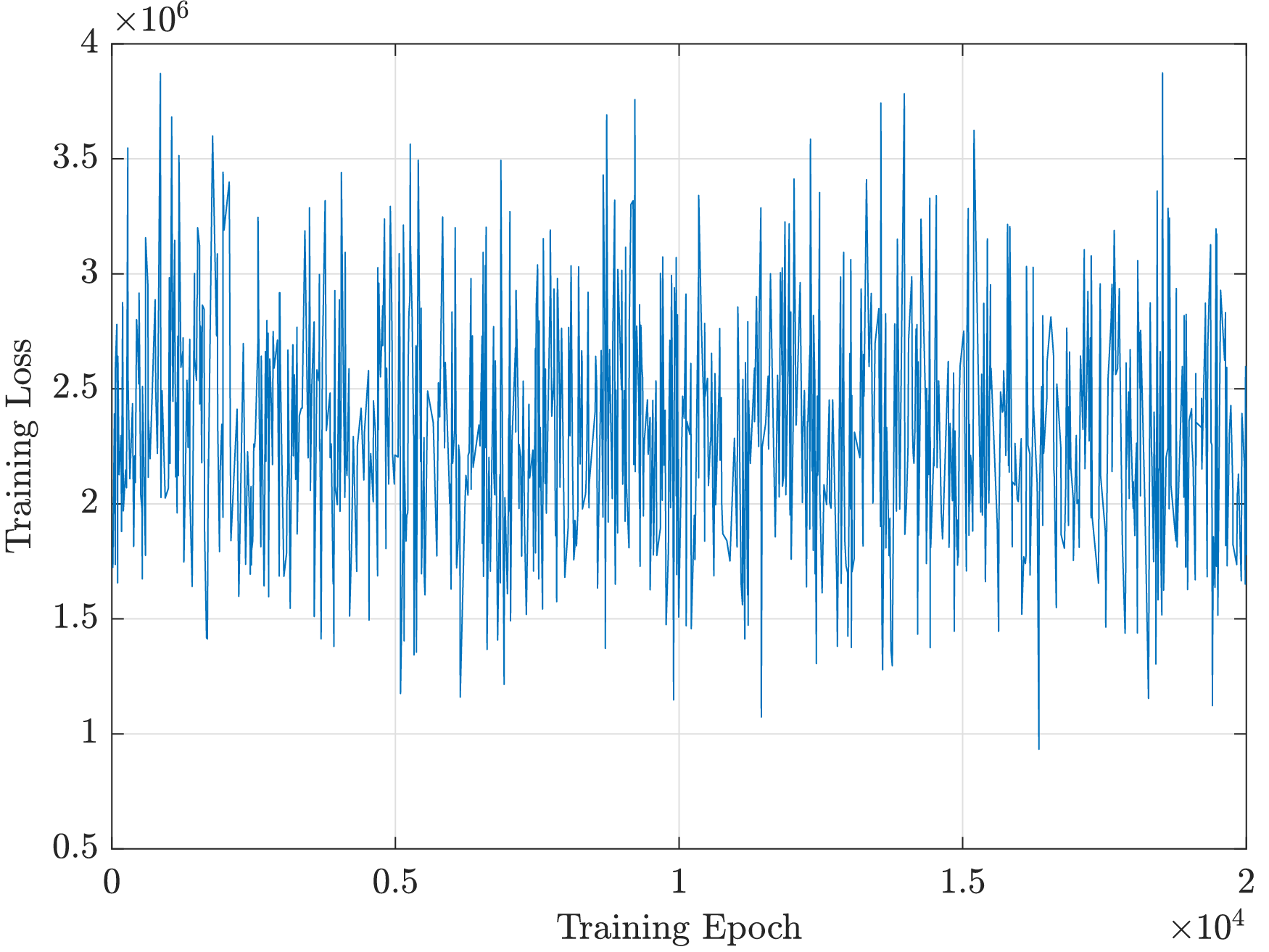}
    \end{subfigure}
    \caption{The 20,000 epoch model using orthogonal arrays (left) and the 20,000 epoch model using random data (right)}
    \label{fig:final}
\end{figure}

To get more precise results, the final ANN architecture underwent training with over 20,000 epochs. The training loss of the ANN model trained on orthogonal dataset is compared to the ANN model trained on random dataset in Figure \ref{fig:final}. All hyperparameters were kept consistent between the two models, and the only difference was the dataset  - one model was using the orthogonal dataset, and the other a dataset of random images. In Figure \ref{fig:final}, it can be observed that the proposed ANN was successfull while training on orthogonal dataset, while the ANN trained on random dataset failed to converge. This can be explained due to the fact that the orthogonal array introduces a structure to the dataset, which can be readily recognized and utilized by the ANN. 

\section{Results}
The proposed ANN was successful in training on the orthogonal database, because the neural network correctly predicted the joint space vector from the training images with an average error of only $2.2$ degrees. These results prove that the ANN was able to learn and converge to a model to base its predictions on. When running inferences on validation database containing random joint position vector with values between -55 degrees to 55 degrees the model was able to estimate the pose within a $21.2$ degree error on average, which is about twice as large as the step that was used to generate the orthogonal training dataset. The prediction had an inaccuracy ranging from $8.6$ degrees to $36.2$ degrees. The results reveal some inconsistencies in the ANN's ability to estimate positions of specific joints, which may explain why the lowest and maximum values differ by such a huge number. Throughout the entire validation dataset, the first and second joint estimations are within $15$ degrees of error. A greater degree of inaccuracy is observed in the third and fourth joints. Inferring from this, it can be assumed that if the network could more accurately estimate the third and fourth joints, the average degree of error would be significantly lower.

The degree of inference inaccuracy was substantially higher in the ANN trained on random data set, indicating that it was unable to learn on the random training data. The average inference error of this model on training database was at $30.3$ degrees which is significantly larger than of the ANN that was trained on the orthogonal database. The validation set was also tested, and the margin of difference between the orthogonal dataset's results and the random dataset's results is much closer, at around $24.1$ degrees. The highest error was 32.4 and the lowest was 18. The random dataset's error deviation is more tight seems then the error spread found in the orthogonal dataset. This could be because the model is so severely underfitted that it only makes one reliable guess that is roughly in the middle of the joint degree limit. 
The underfitting of the ANN assumption is also supported by the large training inference error. In the Table \ref{tab:1} below all joint averages can be seen for both the orthogonal and random dataset on the validation data. 
\begin{table}[h!]

\caption{The inference error in degrees for two ANN models trained on random and orthogonal datasets validated on a randomly generated external dataset of 32 RGB-D images}
\begin{tabular}{||l|llllllll||}\hline
\textbf{Dataset Type} & \textbf{Joint 1} & \textbf{Joint 2} & \textbf{Joint 3} & \textbf{Joint 4} & \textbf{Joint 5} & \textbf{Joint 6} & \textbf{Joint 7} & \textbf{Average} \\  \hline  \hline
Random                & 25.97            & 25.25            & 22.4             & 23.7             & 23.6             & 24.1             & 27.1             & 24.6             \\
Orthogonal            & 6                & 9.2              & 33               & 18.9             & 28               & 25.3             & 37.3             & 21.2 \\ \hline          
\end{tabular}
\label{tab:1}
\end{table}

\begin{figure}
    \centering
    \includegraphics[scale=0.8]{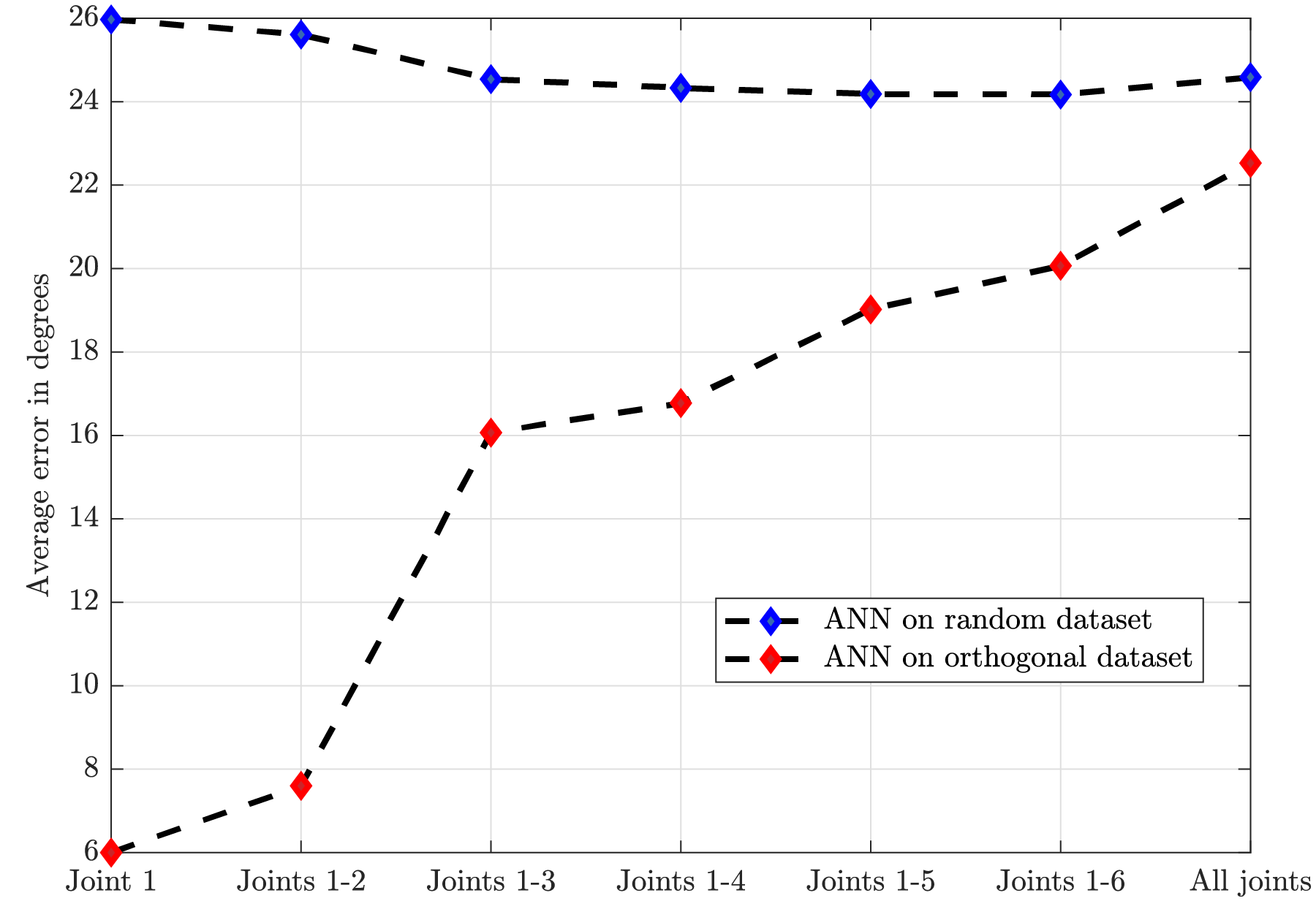}
    \caption{Average inference error on validation dataset for two ANNs trained on random and orthogonal datasets}
    \label{fig:comparsionaans}
\end{figure}

Currently, the inference accuracy of the orthogonal ANN model to a randomly generated external dataset consisting of 32 images is around $6-9^\circ$ for the first two joints and $24-26^\circ$ for the rest of the joints making the average error for all joints to be approximately $24.2^\circ$, which is equal to twice the discretized step formulated using $s = 12$ levels as shown in Equation \ref{eq:orthogonal_array}. The Figure \ref{fig:comparsionaans} demonstrates that the average error for ANN trained on orthogonal dataset jumps from about $8$ degrees to $16$ degrees when first three joints are considered. This demonstrates that it is more challenging for the ANN to accurately estimate joints that are further away from the base of the robotic arm since there are more possibilities for self occlusion.

\section{Conclusion and Future Work}
This work presented a custom ANN architecture that is capable of estimating a joint position vector of a Sawyer robotic manipulator from a single RGB-D image with an average error that is twice the discretized step that was used to generate the training data. In order to efficiently span the $7$ dimensional solution space an orthogonal array method was adopted which significantly reduced the database size from $20$ million to $144$ images. The resulting discretized step is $10^\circ$, whereas the average inference accuracy is about twice as large $(21.2^\circ)$ on ANN trained on orthogonal database validated on $32$ out-of-sample random poses. Interestingly, even with this small orthogonal database the inference accuracy of the first two joints is remarkable at $6^\circ$ and $9.2^\circ$ respectively, which is lower than the discretized step of the database. This finding indicates that the ANN was successful in extrapolating joint position estimates of first two joints given completely random out-of-sample imagery.

The presented ANN architecture was also trained on a control database consisting of $144$ images with random joint position vectors. The training of the ANN on a control database was unsuccessful as the training loss oscillated around a huge number indicating that the model was severely underfit. This is to be expected as training on such a small number of images and mapping the ANN output to a $7$ dimensional solution space is impractical and calls for a structured database which provides guidance and assists with convergence during training. As the training loss did not fall below the stopping criterion the training of ANN on random database was stopped at $20,000$ epochs which is the same number of epochs the final orthogonal based ANN was trained for. The average inference error of the ANN trained on random dataset was consistently above around $24^\circ$, which indicates that this ANN converged to a model that consistently estimates joint positions roughly in the middle of the joint limits. 

In this work the camera angle and position was fixed relative to the base of the robotic arm, in future it is planned to investigate optimal position and point of view for training CV based ANN, which would minimize the self-occlusion. The orthogonal database can be further expanded by linearly interpolating the values in the presented orthogonal array or generate a different orthogonal array with more levels and number of runs. Moreover, it is also planned to improve the inference accuracy by combining the ANN model with forward and inverse kinematics resulting in kinematics informed machine learning model. Furthermore, the training can be further improved by investigating other color spaces (e.g. HSV, LAB, etc.)

\clearpage

\bibliographystyle{AAS_publication}   
\bibliography{AAStemplatev2_0_6}   

\begin{thebibliography}{10}

\bibitem{poe_robotics3}
A.~Malik, T.~Henderson, and R.~Prazenica, ``Multi-Objective Swarm Intelligence
  Trajectory Generation for a 7 Degree of Freedom Robotic Manipulator,''  {\em
  Robotics}, Vol.~10, No.~4, 2021, p.~127.

\bibitem{poe_robotics1}
A.~Malik, T.~Henderson, and R.~J. Prazenica, ``Trajectory generation for a
  multibody robotic system using the product of exponentials formulation,''
  {\em AIAA Scitech 2021 Forum}, 2021, p.~2016.

\bibitem{poe_robotics2}
A.~Malik, Y.~Lischuk, T.~Henderson, and R.~Prazenica, ``Generating Constant
  Screw Axis Trajectories With Quintic Time Scaling For End-Effector Using
  Artificial Neural Network And Machine Learning,''  {\em 2021 IEEE Conference
  on Control Technology and Applications (CCTA)}, IEEE, 2021, pp.~1128--1134.

\bibitem{poe_robotics4}
J.~J. Korczyk, D.~Posada, A.~Malik, and T.~Henderson, ``Modeling of an On-Orbit
  Maintenance Robotic Arm Test-Bed,''  {\em 2021 AAS/AIAA Astrodynamics
  Specialist Conference}, 2021, pp.~August 8--12, Big Sky, USA.

\bibitem{poe_robotics5}
A.~Malik, ``Trajectory Generation for a Multibody Robotic System: Modern
  Methods Based on Product of Exponentials,''  {\em ERAU Commons}, 2021.

\bibitem{poe_robotics6}
A.~Malik, Y.~Lischuk, T.~Henderson, and R.~Prazenica, ``A Deep Reinforcement
  Learning Approach for Inverse Kinematics Solution of a High Degree of Freedom
  Robotic Manipulator,''  {\em Robotics}, Vol.~12, No.~5, 2022, p.~130.

\bibitem{malik2022using}
A.~Malik, T.~Henderson, and R.~J. Prazenica, ``Using Products of Exponentials
  to Define (Draw) Orbits and More,''  {\em Advances in the Astronautical
  Sciences}, Vol.~175, 2021, p.~3319.

\bibitem{poe}
R.~W. Brockett, ``Robotic manipulators and the product of exponentials
  formula,''  {\em Mathematical theory of networks and systems}, Springer,
  1984, pp.~120--129.

\bibitem{lynch2017modern}
K.~M. Lynch and F.~C. Park, {\em Modern robotics}.
\newblock Cambridge University Press, 2017.

\bibitem{pulli2012real}
K.~Pulli, A.~Baksheev, K.~Kornyakov, and V.~Eruhimov, ``Real-time computer
  vision with OpenCV,''  {\em Communications of the ACM}, Vol.~55, No.~6, 2012,
  pp.~61--69.

\bibitem{wasenmuller2016comparison}
O.~Wasenm{\"u}ller and D.~Stricker, ``Comparison of kinect v1 and v2 depth
  images in terms of accuracy and precision,''  {\em Asian Conference on
  Computer Vision}, Springer, 2016, pp.~34--45.

\bibitem{mahendran2021computer}
J.~K. Mahendran, D.~T. Barry, A.~K. Nivedha, and S.~M. Bhandarkar, ``Computer
  vision-based assistance system for the visually impaired using mobile edge
  artificial intelligence,''  {\em Proceedings of the IEEE/CVF Conference on
  Computer Vision and Pattern Recognition}, 2021, pp.~2418--2427.

\bibitem{baruch2021arkitscenes}
G.~Baruch, Z.~Chen, A.~Dehghan, T.~Dimry, Y.~Feigin, P.~Fu, T.~Gebauer,
  B.~Joffe, D.~Kurz, A.~Schwartz, {\em et~al.}, ``ARKitScenes--A Diverse
  Real-World Dataset For 3D Indoor Scene Understanding Using Mobile RGB-D
  Data,''  {\em arXiv preprint arXiv:2111.08897}, 2021.

\bibitem{tome2017lifting}
D.~Tome, C.~Russell, and L.~Agapito, ``Lifting from the deep: Convolutional 3d
  pose estimation from a single image,''  {\em Proceedings of the IEEE
  Conference on Computer Vision and Pattern Recognition}, 2017, pp.~2500--2509.

\bibitem{wang2020deep}
J.~Wang, K.~Sun, T.~Cheng, B.~Jiang, C.~Deng, Y.~Zhao, D.~Liu, Y.~Mu, M.~Tan,
  X.~Wang, {\em et~al.}, ``Deep high-resolution representation learning for
  visual recognition,''  {\em IEEE transactions on pattern analysis and machine
  intelligence}, Vol.~43, No.~10, 2020, pp.~3349--3364.

\bibitem{yu2021lite}
C.~Yu, B.~Xiao, C.~Gao, L.~Yuan, L.~Zhang, N.~Sang, and J.~Wang, ``Lite-hrnet:
  A lightweight high-resolution network,''  {\em Proceedings of the IEEE/CVF
  Conference on Computer Vision and Pattern Recognition}, 2021,
  pp.~10440--10450.

\bibitem{he2019bounding}
Y.~He, C.~Zhu, J.~Wang, M.~Savvides, and X.~Zhang, ``Bounding box regression
  with uncertainty for accurate object detection,''  {\em Proceedings of the
  ieee/cvf conference on computer vision and pattern recognition}, 2019,
  pp.~2888--2897.

\bibitem{zheng2020distance}
Z.~Zheng, P.~Wang, W.~Liu, J.~Li, R.~Ye, and D.~Ren, ``Distance-IoU loss:
  Faster and better learning for bounding box regression,''  {\em Proceedings
  of the AAAI Conference on Artificial Intelligence}, Vol.~34, 2020,
  pp.~12993--13000.

\bibitem{rezatofighi2019generalized}
H.~Rezatofighi, N.~Tsoi, J.~Gwak, A.~Sadeghian, I.~Reid, and S.~Savarese,
  ``Generalized intersection over union: A metric and a loss for bounding box
  regression,''  {\em Proceedings of the IEEE/CVF conference on computer vision
  and pattern recognition}, 2019, pp.~658--666.

\bibitem{zhang2007orthogonal}
Y.~Zhang, ``Orthogonal arrays obtained by repeating-column difference
  matrices,''  {\em Discrete Mathematics}, Vol.~307, No.~2, 2007, pp.~246--261.

\bibitem{hedayat1999orthogonal}
A.~S. Hedayat, N.~J.~A. Sloane, and J.~Stufken, {\em Orthogonal arrays: theory
  and applications}.
\newblock Springer Science \& Business Media, 1999.

\bibitem{roy2010primer}
R.~K. Roy, {\em A primer on the Taguchi method}.
\newblock Society of Manufacturing Engineers, 2010.

\bibitem{LeslieHyp}
L.~N. Smith, ``A disciplined approach to neural network hyper-parameters: Part
  1 -- learning rate, batch size, momentum, and weight decay,''  2018,
  10.48550/ARXIV.1803.09820.

\end{thebibliography}

\end{document}